\documentclass{spie}

\usepackage{cite}

\usepackage{times} 
\usepackage{indentfirst} 

\usepackage{amsmath} 

\usepackage{amsfonts}

\def\ZZ{\mathbb{Z}}

\usepackage{multirow}

\usepackage[colorinlistoftodos,color=orange]{todonotes} 
\usepackage{xkeyval}
\presetkeys{todonotes}{inline}{}

\title{The Analysis of Projective Transformation Algorithms for Image Recognition on Mobile Devices}

\author{Anton Trusov\textsuperscript{1, 3}, Elena Limonova \textsuperscript{2, 3, 4}
  \skiplinehalf
  \normalsize 
  \textsuperscript{1} Moscow Institute of Physics and Technology, Dolgoprudnii, Russia; \\
  \textsuperscript{2} Institute for Systems Analysis, FRC CSC RAS, Moscow, Russia; \\
  \textsuperscript{3} Smart Engines Service LLC, Moscow, Russia;\\
  \textsuperscript{4} Institute for Information Transmission Problems RAS, Moscow, Russia;   
}

\begin{document}

\maketitle

\begin{abstract}
  In this work we apply commonly known methods of non-adaptive interpolation (nearest pixel, bilinear, B-spline, bicubic, Hermite spline) and sampling (point sampling, supersampling, mip-map pre-filtering, rip-map pre-filtering and FAST) to the problem of projective image transformation. We compare their computational complexity, describe their artifacts and than experimentally measure their quality and working time on mobile processor with ARM architecture. Those methods were widely developed in the 90s and early 2000s, but were not in an area of active research in resent years due to a lower need in computationally efficient algorithms. However, real-time mobile recognition systems, which collect more and more attention, do not only require fast projective transform methods, but also demand high quality images without artifacts. As a result, in this work we choose methods appropriate for those systems, which allow to avoid artifacts, while preserving low computational complexity. Based on the experimental results for our setting they are bilinear interpolation combined with either mip-map pre-filtering or FAST sampling, but could be modified for specific use cases.

\end{abstract}

\section{Introduction}

Image recognition systems have become wide spread in resent years. They are used in many application areas including mobile devices and embedded systems. Practical usage on different kinds of hardware imposes some constrains on algorithms for those applications. For example, algorithms with high computational complexity can not be used in real-time, however, decrease of computational complexity in many cases leads to poor recognition quality. Moreover, mobile devices do not have as powerful graphics processing units (GPUs) as those of work stations or specialized servers. Also it is hard to develop cross-platform applications for GPUs because of the lack of standard tools supporting various devices. Sometimes data transfer from central processing unit (CPU) to GPU and back may take quite a while compared to computation time. All those factors force us to conduct calculations on CPU for mobile devices. To increase computational performance we can use availiable CPU cores and SIMD extensions, if they are supported by the architecture \cite{limonova2015, limonova2016}.

One of basic operations of image processing is projective transformation. It is used in the number of image recognition systems for rectification of input image from camera (for example in real-time document recognition systems \cite{bulatov2018smart}, real-time recognition of TV content \cite{skoryukina2017snapscreen}, or visual navigation of unmanned aerial vehicles \cite{stepanov2015}). Another application area of projective transformation is rendering of 3D scenes in computer graphics \cite{heckbert1986survey} in case they are not rendered by computationally complex ray tracing algorithm \cite{maltsev2012}. This area is widely described in literature, however, authors of such algorithms do not usually pay much attention to computational efficiency and implementation aspects important for mobile devices.

It is also worth noting, that there are many methods of interpolation and anti-aliasing used in projective transformation. It happens due to the fact that some parts of image may be scaled up or down as a result of transformation. In up-scaled parts interpolation is needed to get value of signal "between" pixels of the source image. In down-scaled parts some actions need to be taken to avoid aliasing, which can cause artifacts such as moire pattern.

In this article we look into different variations of projective image transformation, which use common methods of interpolation and anti-aliasing. Most of those methods are well known and have been developed in 90s or early 2000s however became forgotten in sphere of recognition systems in recent years due to increase in computational power, but problems of computational efficiency remain crutial for mobile applications. We compare computational complexity of these algorithms and their quality and provide experimental time measurements on mobile CPU with ARM architecture. The results allow us to choose algorithm of projective transformation that fits constraints of specific task on mobile device.

\section{Projective image transformation}

Let us consider how digital image of planar object is formed by camera. Digital image is a result of sampling of optical image, formed by camera. In case of simple camera models (for example, camera obscura or pinhole camera) is just a central projection of planar object. Since, optical images of the planar object formed in two different points of view are both central projections, they are connected by projective transformation of a plane, so if we need an image as if it was taken from a different view point (e.g. orthogonal projection of a document), we can compute it using projective transformation of an image we already have. Projective transformation in Cartesian coordinate system can be expressed as: (\ref{eq_projective}).

\begin{equation}
    \begin{split}
        x =  \frac{t_{11} u + t_{12} v + t_{13}}{t_{31} u + t_{32} v + t_{33}}; \\
        y =  \frac{t_{21} u + t_{22} v + t_{23}}{t_{31} u + t_{32} v + t_{33}},
        \label{eq_projective}
    \end{split}
\end{equation}

%

 
 where $T = (t_{ij}), i \in \{1, 2, 3\}, j \in \{1, 2, 3\} $ is the matrix of projective transformation, $(u, v)$ and $(x, y)$ are the coordinates on planes of original and transformed images.
 
 However, we can not apply (\ref{eq_projective}) directly to compute projective transformation of one digital image into another, because both input and output images are discrete. According to \cite{wolberg1990} projective transformation of one digital image is special case of image resampling problem, which consists of 3 stages:
 
 \begin{enumerate}
    \item \textbf{Image reconstruction.} During this stage we retrieve continuous approximation of optical image $\tilde{I}(u, v)$ from digital image $I_{m \times n}$, where ${m \times n}$ is image size.
    \item \textbf{Spatial transformation.} On this stage we apply spatial transformation, which is in our case projective transformation (\ref{eq_projective}), to get optical image from another point of view $\tilde{J}(x(u, v), y(u, v)) = \tilde{I}(u, v)$.
    \item \textbf{Image sampling.} Finally, we sample optical image $\tilde{J}(x, y)$ to get digital one $J_{k\times l}$.
 \end{enumerate}
 
 However, these steps are not an algorithm, because second stage can not be directly computed using finite number of operations. That is why we consider algorithms based on inverse mapping:

\begin{enumerate}
    \item We choose coordinates of sampling points on the output image, probably more than one per pixel (sampling grid).
    \item For each point we calculate its coordinates on input image using inverse transformation of (\ref{eq_projective}).
    \item For each point we calculate value of it's intensity using selected interpolation of nearby pixels of input image.
    \item Finally, we compute values of pixels of output image from sampling points.
 \end{enumerate}

In these algorithms we can choose different interpolation functions and ways of selection and weighting of sampling points.

\subsection{Interpolation}

Interpolation is a way to get values of image intensity "between" pixels of source image. It is necessity becomes obvious when we deal with image magnification and as projective transformation can lead to magnification of some image parts we need interpolation in our algorithm.

If we assume usage of simple image sampling model, which defines value of a pixel as an intensity of an optical image in its center, we can in theory derive ideal restoration method. This method is a convolution with ideal low pass filter (LPF), and it will restore optical image precisely, if Nyquist frequency (half of a sampling frequency) is lower than maximal frequency in Fourier spectrum of an optical image \cite{diniz2010digital}. In fact this is not a real-life scenario for several reasons: firstly, sampling in real optical systems is not a point sampling and, which is more important, in real objects have sharp borders and therefore image has infinite Fourier spectrum and ideal restoration is not possible.

So, we consider kernel-based interpolation functions (\ref{eq_inter}). There are different approaches to image reconstruction, for example adaptive interpolation based on local gradient features \cite{hwang2004adaptive} or local isophote orientation \cite{wang2007new}. However, they are more computationally complex than non-adaptive methods, so we do not consider them in this work. In all examples we provide one-dimensional interpolation
\begin{equation}
    \tilde{I}(x) = \sum_{k \in \ZZ} h(x - k) I_k,
    \label{eq_inter}
\end{equation}

where $I_k$ define pixels of the input image, $h(x)$ is interpolation kernel.

\begin{figure}[ht]
    \begin{minipage}[h]{0.245\linewidth}
    \center{\includegraphics[width=1\linewidth]{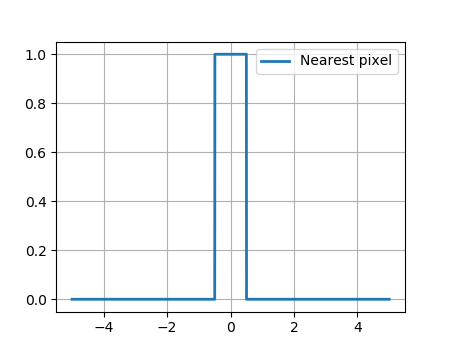} \\
            \includegraphics[width=1\linewidth]{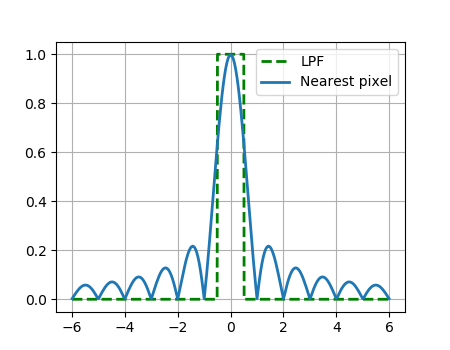} \\a)}
    \end{minipage}
    \hfill
    \begin{minipage}[h]{0.245\linewidth}
    \center{\includegraphics[width=1\linewidth]{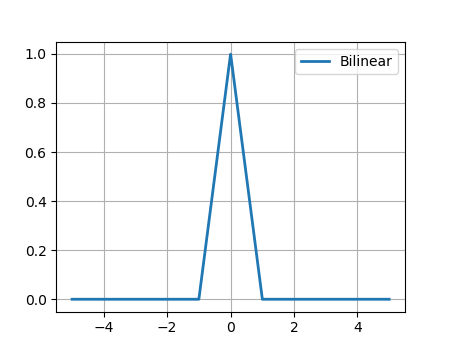} \\
            \includegraphics[width=1\linewidth]{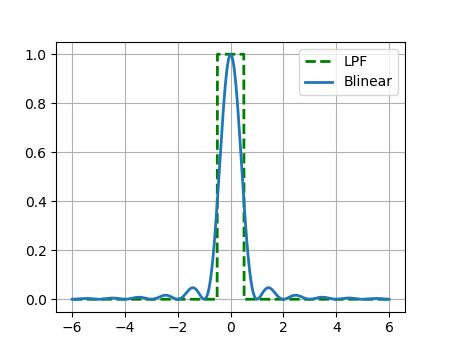} \\b)}
    \end{minipage}            
    \hfill
    \begin{minipage}[h]{0.245\linewidth}
    \center{\includegraphics[width=1\linewidth]{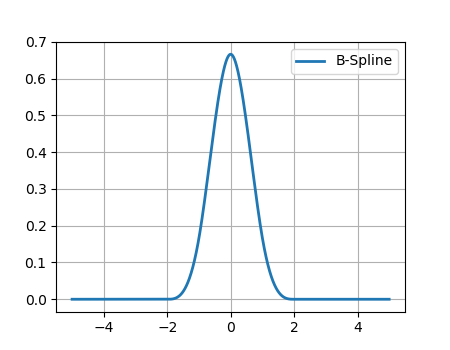} \\
            \includegraphics[width=1\linewidth]{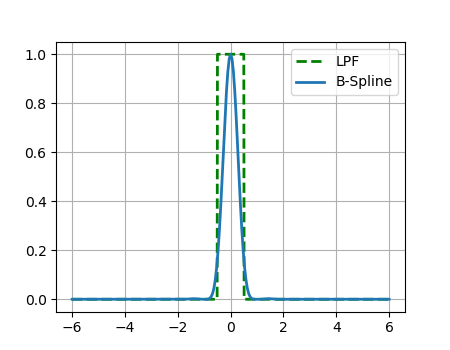} \\c)}
    \end{minipage}
    \hfill
    \begin{minipage}[h]{0.245\linewidth}
    \center{\includegraphics[width=1\linewidth]{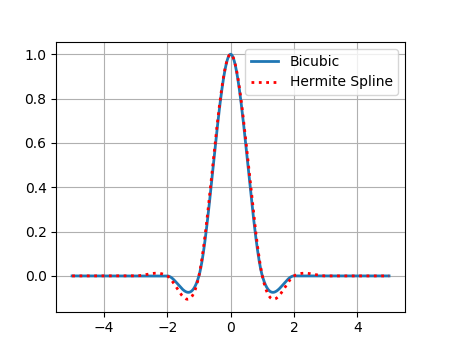} \\
            \includegraphics[width=1\linewidth]{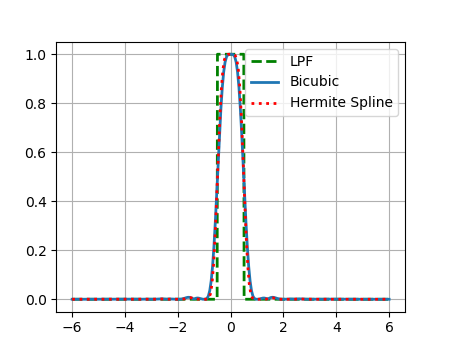} \\d)}
    \end{minipage}      
    \caption{Interpolation kernels (fist row) and Fourier spectrum (second row). a) Nearest pixel, b) Bilinear, c) B-spline, d) Bicubic and Hermite spline}
    \label{fig_inter}
\end{figure}

\textbf{Nearest-pixel interpolation} is the most computationally simple interpolation, which uses box-filter (\ref{eq_i0}). Its Fourier spectrum is very different from that of LPF (Fig. \ref{fig_inter}a). 
\begin{equation}
    h(x) =
    \begin{cases}
    1 , \mbox{  } |x| < 0.5\\
    0 , \mbox{  } |x| \geq 0.5
    \end{cases}
    \label{eq_i0}
\end{equation}

\textbf{Bilinear interpolation} is two-dimensional analog of linear interpolation. Its kernel (\ref{eq_i1}) is continuous therefore optical image, reconstructed by this kernel, is also continuous (Fig. \ref{fig_inter}b). It works good enough far from edges, but smooths object edges.
\begin{equation}
    h(x) =
    \begin{cases}
    1 - |x| , \mbox{  } |x| < 1\\
    0 , \mbox{ } |x| \geq 1
    \end{cases}
    \label{eq_i1}
\end{equation}

\textbf{Bicubic interpolation} is two-dimensional analog of linear cubic interpolation. Its kernel has continuous first derivative and so has optical image, reconstructed by this kernel (\ref{eq_i3}) (Fig. \ref{fig_inter}d). Parameter $\alpha$ is usually set to $-0.5$ \cite{keys1981cubic}. Far from edges this interpolation produces smooth image that looks fine, however, artifact known as halo appears on the edges (bright line alongside brighter side of the edge and dark line alongside darker).
\begin{equation}
h(s) = 
    \begin{cases}
    (2 + \alpha)|s|^3 - (3 + \alpha)|s|^2 + 1, \mbox{  }  |s| < 1 \\
    \alpha|s|^3 - 5 \alpha|s|^2 + 8 \alpha|s| - 4 \alpha, \mbox{  }  1 \le |s| < 2 \\
    0, \mbox{  } |s| \geq 2
    \end{cases}
\label{eq_i3}
\end{equation}

To get rid of halo we need kernel to be positive. We also want it to be smooth and have small support (so that we use as small number of pixels as possible for computation of interpolated value). Kernel of \textbf{cubic B-spline} (\ref{eq_ibs}) satisfies these conditions. Its Fourier spectrum is very thin (Fig. \ref{fig_inter}c), that means reconstructed image is over-smoothed, which however can be a good thing, if the input image is nosy.
\begin{equation}
h(s) = \frac{1}{6}
    \begin{cases}
    3|s|^3 - 6|s|^2 + 4, \mbox{  }  |s| < 1 \\
    -|s|^3 + 6|s|^2 -12|s| + 8, \mbox{  }  1 \le |s| < 2 \\
    0, \mbox{  } |s| > 2
    \end{cases}
    \label{eq_ibs}
\end{equation}

\textbf{Cubic Hermite spline} is yet another interpolation method. If we define Hermit polynomials as:
\begin{equation}
    \begin{split}
    &h_{00}(t) = 2t^3 - 3t^2 + 1, \\
    &h_{01}(t) = -2t^3 + 3t^2,  \\
    &h_{10}(t) = t^3 - 2t^2 + t, \\
    &h_{11}(t) = t^3 - t^2,
    \end{split}
\end{equation}
than interpolation function on between 0 and 1 will be:
 \begin{equation}
          g(x) = f(0) h_{00}(x) + f(1) h_{01}(x) + f'(0) h_{10}(x) + f'(1) h_{11}(x),
    \label{eq_hs_inter}
\end{equation}
 where $f(.)$ are values of target function we want to reconstruct and $f'(.)$ are its derivatives. It may seem that this is not kernel interpolation, however, if we select fixed difference approximation for derivative, it will become one. For example, if we approximate derivative as:
 \begin{equation}
        \begin{split}
          f'(n) &= \frac{f(n + 1) - f(n -1)}{2} 
        \end{split}
\end{equation}

we will get cubic kernel with $\alpha = -0.5$. So, to get a different kernel and to increase quality of interpolation we choose  \cite{seta2009digital}:
\begin{equation}
        \begin{split}
          f'(n) &= -\frac{1}{12}f(n + 2) + \frac{2}{3}f(n + 1) - \frac{2}{3}f(n - 1) + \frac{1}{12}f(n - 2)
        \end{split}
        \label{deriv_2}
\end{equation}
resulting filter's Fourier spectrum is slightly closer to that of LPF, than spectrum of cubic kernel (Fig. \ref{fig_inter}d). This interpolation method still leads to halo artifact. 

\begin{figure}[ht]
    \begin{minipage}[h]{0.15\linewidth}
    \center{\includegraphics[width=1\linewidth]{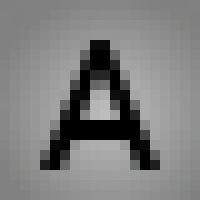} \\ a)}
    \end{minipage}
    \hfill
    \begin{minipage}[h]{0.15\linewidth}
    \center{\includegraphics[width=1\linewidth]{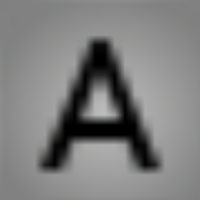} \\ b)}
    \end{minipage}            
    \hfill
    \begin{minipage}[h]{0.15\linewidth}
    \center{\includegraphics[width=1\linewidth]{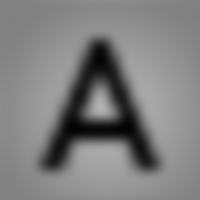} \\ c)}
    \end{minipage}
    \hfill
    \begin{minipage}[h]{0.15\linewidth}
    \center{\includegraphics[width=1\linewidth]{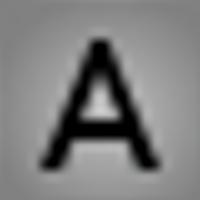} \\ d)}
    \end{minipage}            
    \hfill
    \begin{minipage}[h]{0.15\linewidth}
    \center{\includegraphics[width=1\linewidth]{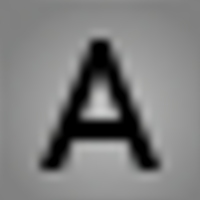} \\ e)}
    \end{minipage} 
    \caption{Interpolation examples. a) Nearest pixel, b) Bilinear, c) B-spline, d) Bicubic, e) Hermite spline}
    \label{fig_inter_ex}
\end{figure}

Computational complexity of all the above methods is defined by the number of multiplications and number of memory accesses (number of pixels of input image, needed for interpolation of one point) per pixel point. We can decrease the number of multiplications to number of memory accesses, if we approximate kernel by piece-wise constant function, and save it into pre-calculated table \cite{feibush1980synthetic}. In table \ref{tab_inter} we represent computational complexity of methods described above. In Fig. \ref{fig_inter_ex} there are an examples of character "A", which are up-scaled 10 times using above interpolation methods.

\begin{table}[ht]
    \caption{Computational complexity of interpolation methods.}
    \label{tab_inter}
    \begin{center}
        \begin{tabular}{|l|c|c|}
        \hline
        \textbf{Interpolation} & \textbf{Multiplications per point} & \textbf{Memory accesses per point}\\
        \hline
        Nearest pixel & 0 & 1\\
        \hline
        Bilinear & 8 & 4\\
        \hline
        Bicubic & 68 & 16\\
        \hline
        Cubic B-spline & 68 & 16\\
        \hline
        Cubic Hermite spline& 76 & 36\\
        \hline
        \end{tabular}
    \end{center}
\end{table}

\subsection{Sampling}

We consider that digital image $I_{M \times N}$ is computed from optical $I(u, v)$ through convolution with sampling filter $h_s(x, y)$ (\ref{eq_filter}).

\begin{equation}
    I_{ij} = \iint I(u, v)h_s(i-u, j - v)du~dv,
    \label{eq_filter}
\end{equation}

Though this expression can not be calculated directly and sampling filter is defined by physical properties of optical system, we can approximate it using finite number of sampling points and some model of a filter. However, if sampling rate is too low, there will be artifacts on the resulting image caused by aliasing of Fourier spectra, because conditions of Nyquist–Shannon sampling theorem are not met. So, once again we need to balance between the computational complexity and the quality of an output image. 

The simplest sampling method is \textbf{point sampling}, where $h_s = \delta(u, v)$. We simply calculate value in center of every pixel of the output image. However it leads to the most significant aliasing artifacts.

To avoid those artifacts we can calculate value using sampling grid of smaller step and then take average over pixel. This method is known as \textbf{supersamplig} and uses $h_s$ equal to a box-filter. Choosing different $h_s$ will lead to similar methods, which will nevertheless have significantly higher computational complexity, than point-sampling.

One may notice that aliasing artifacts appear as a result of undersampling in down-scaled parts of image, so to avoid them we can downscale input image, and then apply point sampling. This process is called \textbf{pre-filtering}. A way to implement it is using a \textbf{mip-map} \cite{williams1983pyramidal}, which is an image pyramid where an image on each next level in twice down-scaled copy of a previous image. For each pixel of output image we estimate local scale factor (how significantly this part of image is down-scaled) and choose mip-map level accordingly. This estimation can be done as follows \cite{ewins1998mip}. The scale factor is $\max \left (\sqrt{\Big(\frac{\partial u}{\partial x} \Big) ^2 + \Big(\frac{\partial v}{\partial x} \Big) ^2}, \sqrt{\Big(\frac{\partial u}{\partial y} \Big) ^2 + \Big(\frac{\partial v}{\partial y} \Big) ^2} \right) $, where $x$ and $y$ are coordinates on the output image, $u$ and $v$ are coordinates on the input one. To avoid sharp borders between regions, where different levels of mip-map are selected, we use linear interpolation between two nearest mip-map layers. 

Mip-map approximates an area of output pixel on the input image (which is either quadrangle, if we consider rectangular $h_s$, or ellipse, if $h_s$ has radial symmetry) by a square, which is incorrect in general case, especially if projection is highly anisotropic. To partly overcome this problem we can use data structure like mip-map, which scales image along its height and width independently \cite{heckbert1989fundamentals}. It is known as \textbf{rip-map}. Rip-map approximates pixel area as rectangular with sides parallel to coordinate axis of input image. However, the direction of anisotropy is not necessary parallel to this direction.

To overcome this problem, we can combine ideas of mip-map pre-filtering and supersamplig. Authors of algorithm \textbf{FAST} \cite{chen2004footprint} proposed to use jittered sampling over mip-map level, defined by minor axis of anisotropy, where the number of samples is defined by major axis.

\begin{figure}[ht]
    \begin{minipage}[h]{0.17\linewidth}
    \center{\includegraphics[width=1\linewidth]{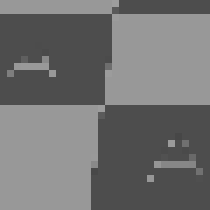} \\ a)}
    \end{minipage}
    \hfill
    \begin{minipage}[h]{0.17\linewidth}
    \center{\includegraphics[width=1\linewidth]{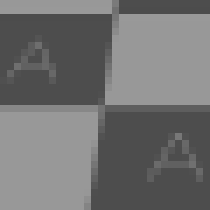} \\ b)}
    \end{minipage}            
    \hfill
    \begin{minipage}[h]{0.17\linewidth}
    \center{\includegraphics[width=1\linewidth]{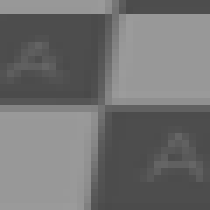} \\ c)}
    \end{minipage}
    \hfill
    \begin{minipage}[h]{0.17\linewidth}
    \center{\includegraphics[width=1\linewidth]{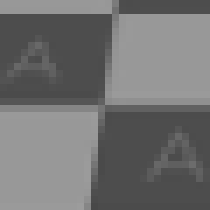} \\ d)}
    \end{minipage}            
    \hfill
    \begin{minipage}[h]{0.17\linewidth}
    \center{\includegraphics[width=1\linewidth]{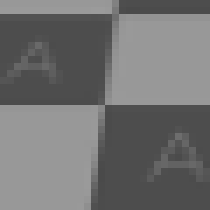} \\ e)}
    \end{minipage} 
    \caption{Sampling examples. a) Point sampling, b) Supersampling, c) Mip-map pre-filtering, d) Rip-map pre-filtering, e) FAST}
    \label{fig_samp}
\end{figure}

Computational complexity of sampling methods is defined by number of samples per pixel and number of multiplications per sample. If we decompose backward projection as:
\begin{equation}
        \begin{split}
            q(x, y) &= \frac{1}{f_3(x) + g_3(y)} \\
            u(x, y) &=  q(x, y)(f_1(x) + g_1(y)) \\
            v(x, y) &=  q(x, y)(f_3(x) + g_3(y)),
            \label{eq4}
        \end{split}
\end{equation}

where $f_i(x) = p_{i1} x, g_i(y) = p_{i2} y + p_{i3}, P = (p_{ij}) = T ^ {-1}, i \in \{1, 2, 3\}, j \in \{1, 2, 3\} $, $P$ is matrix of backward projection, we will need 2 multiplications and 1 division per sample of uniform sampling grid, because $f_i(x)$ and $g_i(y)$ can be precomputed. Calculating of partial derivatives, rescaling and interpolation between levels for mip-map and rip-map will add 5 multiplications per sample (derivatives can be decomposed as well). FAST sampling may require a lot of samples per pixel in case of significant anisotropy, so their number is usually limited with some constant $n$. Also mip-map and rip-map require additional memory for data structure. In the Table \ref{tab_aa} we represent computational complexity of the methods described above. In the Figure \ref{fig_samp} the examples of the methods are shown in process of projective transformation of letter "A".

 \begin{table}[ht]
    \caption{Computational complexity of sampling methods.}
    \label{tab_aa}
    \begin{center}
    \begin{tabular}{|l|c|c|c|}
        \hline
        \textbf{Sampling} & \textbf{Samples per pixel} & \textbf{Multiplications per sample} & \textbf{Aditional memory}\\
                       &  & & \textbf{(in image size)}\\
        \hline
        Point sampling & 1 & 2  & 0 \\
        \hline
        Supersampling & $n$ & 2 & 0\\
        \hline
        Mip-map pre-filtering & 2 & 7 & $1/3$\\
        \hline
        Rip-map pre-filtering& 4 & 7 & 3\\
        \hline
        FAST & $\leq n$ & 7 & $1/3$\\
        \hline
    \end{tabular}
    \end{center}
\end{table}

\section{Experiments}

To experimentally estimate computational complexity and quality of different algorithms of projective image transformation, we apply a set of 3 projective transformations to an input image, so that their composition is equal transformation, then measure average time and average quality metric PSNR (\ref{eq_psnr}) between the input $S_{K \times L}$ and the output $J_{K \times L}$ images. 

\begin{equation}
    \begin{split}
        &PSNR(J_{K \times L}, S_{K \times L}) = 10 \log_{10} \left( \frac{K L\left (\max \limits_{i, j}(S_{i,j}) \right)^2}{\sum_{i = 0}^{K - 1}\sum_{j = 0}^{L - 1} (J_{ij} - S_{ij})^2}\right).    \end{split}
    \label{eq_psnr}
\end{equation}

To conduct experiment we took 20 document images of size $512\times512$ pixels and 10 matrix sets and measured time and quality on microcomputer ODROID-XU4 with ARM Cortex-A7 CPU (2000 MHz and 2GB of RAM). The results of the experiment are represented in Table \ref{tab_res}.

Experiments confirm that usage of \textit{nearest pixel} as interpolation or \textit{point sampling} as sampling method leads to pure quality, \textit{Hermite spline} interpolation and \textit{supersampling} are too computationally complex and take much time to apply. This makes those methods ineffective in practical real-time applications for mobile devices. It worth mentioning, that \textit{mip-map pre-filtering} and \textit{cubic B-spline} interpolation lead to image blure and as a result to low PSNR, however, in some cases that may not be a serious problem. Usage of \textit{bicubic} and \textit{cubic B-spline} interpolation with any sampling method but \textit{point sampling} also leads to high computational complexity and working time. \textit{Rip-map pre-filtering} requires a lot of additional memory and at the same time does not work as good as \textit{FAST}, while having approximately the same computational complexity and time of work. So, the best choice for real-time applications on mobile devices seems to be algorithms of image projective transformation based on combination of \textit{bilinear} interpolation with either \textit{mip-map} or \textit{FAST} sampling. The first works faster, but the letter provides better quality.

\begin{table}[ht]
    \caption{Time and quality measurements for considered interpolation and sampling methods.}
    \label{tab_res}
    \begin{center}
    \begin{tabular}{|c|c|c|c|}
    \hline
    \textbf{Sampling} & \textbf{Interpolation} & \textbf{Time (sec)} &  \textbf{PSNR (dB)}\\
    
    \hline
    \multirow{5}{*}{Point sampling} 
        &Nearest pixel  & 0.074 & 23.1 \\
        \cline{2-4}
        &Bilinear  & 0.161 & 25.4 \\
        \cline{2-4}
        &Bicubic & 0.496 & 26.0 \\
        \cline{2-4}
        &Cubic B-spline & 0.517 & 24.8 \\
        \cline{2-4}
        &Hermite spline & 0.978 & 26.2 \\
        \cline{2-4}
    
    \hline
    \multirow{5}{*}{Supersampling (49 samples)} 
        &Nearest pixelе & 4.01 & 25.6 \\
        \cline{2-4}
        &Bilinear & 7.76 & 25.3 \\
        \cline{2-4}
        &Bicubic & 23.7 & 26.1 \\
        \cline{2-4}
        &Cubic B-spline & 24.7 & 24.6 \\
        \cline{2-4}
        &Hermite spline & 46.9 & 26.2 \\
        \cline{2-4}
        
    \hline
    \multirow{5}{*}{Mip-map pre-filtering} 
        &Nearest pixel & 0.194 & 24.0 \\
        \cline{2-4}
        &Bilinear & 0.302 & 24.6 \\
        \cline{2-4}
        &Bicubic & 0.770  & 25.3 \\
        \cline{2-4}
        &Cubic B-spline & 0.800  & 23.9 \\
        \cline{2-4}
        &Hermite spline & 1.56 & 25.5\\
        \cline{2-4}
        
    \hline
    \multirow{5}{*}{Rip-map pre-filterin} 
        &Nearest pixel & 0.328 & 24.2 \\
        \cline{2-4}
        &Bilinear & 0.510 & 25.0 \\
        \cline{2-4}
        &Bicubic & 1.231 & 25.8 \\
        \cline{2-4}
        &Cubic B-spline &  1.270  & 24.2 \\
        \cline{2-4}
        &Hermite spline & 2.41 & 25.8 \\
        \cline{2-4}
        
    \hline
    \multirow{5}{*}{FAST} 
        &Nearest pixel & 0.342 & 24.0 \\
        \cline{2-4}
        &Bilinear & 0.539 & 25.2 \\
        \cline{2-4}
        &Bicubic & 1.31 & 26.0 \\
        \cline{2-4}
        &Cubic B-spline & 1.36 & 24.4 \\
        \cline{2-4}
        &Hermite spline & 2.42 & 26.2 \\
        \cline{2-4}
    \hline
    \end{tabular}
    \end{center}
\end{table}

\section{Discussion \& Conclusion}
In this work we compared different algorithms of projective transformation of digital image that use different interpolation methods: nearest pixel, bilinear, B-spline, bicubic, Hermite spline and different sampling techniques: point sampling, supersampling, mip-map pre-filtering, rip-map pre-filtering and FAST. Estimation of computational complexity and experiments on typical mobile CPU confirmed that bicubic, B-spline and especially Hermite spline interpolations are too hard to compute in real-time mobile applications, while quality of biliner interpolation is higher than that of nearest pixel interpolation. At the same time supersampling is also too computationally complex, rip-map pre-filtering requires more memory, while its quality is lower than that of FAST sampling and point sampling produces significant aliasing artifacts. According to those results best choices or real-time applications are algorithms of projective image transformation that use bilinear interpolation combined with either mip-map pre-filtering or FAST sampling.

It worth mentioning, that if application does not have real-time constraint one might prefer bicubic, B-spine or one of adaptive interpolation methods to increase quality of projective transformation. 

\acknowledgements

This work is partially supported by Russian Foundation for Basic Research (projects 17-29-03297, 18-07-01387). 

\bibliographystyle{spiebib}
\bibliography{main}

\end{document}